\documentclass[runningheads]{llncs}
\usepackage{amssymb}
\usepackage{booktabs}
\usepackage{multirow}
\usepackage{graphicx}
\usepackage{pdflscape}
\usepackage{amsmath}
\usepackage{amsfonts}
\usepackage{xcolor}
\usepackage{url}
\usepackage{hyperref}
\usepackage{rotating}
\usepackage{float}

\begin{document}
\setlength{\abovedisplayskip}{10pt}
\setlength{\belowdisplayskip}{10pt}
\title{An effect analysis of the balancing techniques on the counterfactual explanations of student success prediction models}
\titlerunning{Counterfactual explanations of student success prediction models}
%
\author{\href{https://orcid.org/0000-0002-6172-5449}{\includegraphics[scale=0.06]{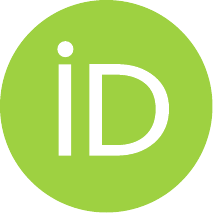}\hspace{1mm}Mustafa Cavus}\inst{1\thanks{corresponding author}} \and
\href{https://orcid.org/0000-0002-8656-0599}{\includegraphics[scale=0.06]{orcid.pdf}\hspace{1mm}Jakub Kuzilek}\inst{2}}
%
\authorrunning{Cavus and Kuzilek}
\institute{Eskisehir Technical University, Department of Statistics, Turkiye \\
\email{mustafacavus@eskisehir.edu.tr} \and
Humboldt University of Berlin, Unter den Linden
6, Berlin, Germany\\
\email{jakub.kuzilek@hu-berlin.de}}
%
\maketitle              

\begin{abstract}
In the past decade, we have experienced a massive boom in the usage of digital solutions in higher education. Due to this boom, large amounts of data have enabled advanced data analysis methods to support learners and examine learning processes. One of the dominant research directions in learning analytics is predictive modeling of learners' success using various machine learning methods. To build learners' and teachers' trust in such methods and systems, exploring the methods and methodologies that enable relevant stakeholders to deeply understand the underlying machine-learning models is necessary. In this context, counterfactual explanations from explainable machine learning tools are promising. Several counterfactual generation methods hold much promise, but the features must be actionable and causal to be effective. Thus, obtaining which counterfactual generation method suits the student success prediction models in terms of desiderata, stability, and robustness is essential. Although a few studies have been published in recent years on the use of counterfactual explanations in educational sciences, they have yet to discuss which counterfactual generation method is more suitable for this problem. This paper analyzed the effectiveness of commonly used counterfactual generation methods, such as WhatIf Counterfactual Explanations, Multi-Objective Counterfactual Explanations, and Nearest Instance Counterfactual Explanations after balancing. This contribution presents a case study using the Open University Learning Analytics dataset to demonstrate the practical usefulness of counterfactual explanations. The results illustrate the method's effectiveness and describe concrete steps that could be taken to alter the model's prediction.

\end{abstract}

\newpage
\section{Introduction}

For centuries universities have been collecting information about their students. With the rise of Information and Communication Technologies, the information collected is transformed from paper-based collections to digital domains. The introduction of new digital education formats and the information collection shift resulted in storing vast amounts of student and study-related data including student demographics, assessment, learning design, and context. In combination with the advancement in Data Mining and Machine Learning (ML) research, the collected data enabled new research exploring the educational domain. The most prominent research fields are Educational Data Mining (EDM) and Learning Analytics (LA), which explore the educational domain from two different perspectives (Siemens and Baker, 2012). More recently, the concerns about the use of Artificial Intelligence (AI) have become stronger uncovering the limitations and possible problems such as bias and explainability of models developed. As a consequence, new data and AI regulations such as GDPR and the AI Act in the EU have been established (Hoel et al., 2017). As a consequence trust in the analytical tools and AI methods in higher education has been reduced leading to the new approach in LA research called Trusted Learning Analytics (TLA) (Drachsler H., 2018). The TLA approach focuses on using intrinsically explainable `white box` AI models and systems. This significantly reduces the opportunity of using more “user-unfriendly” models such as Random Forests (RF) or Neural Networks. Luckily, the field of Explainable Artificial Intelligence (XAI) provides researchers with the methods with the potential to unlock the `black box` models for use in TLA systems (Drachsler H., 2018). The trend of using XAI methods in the educational domain is highly resonating within the research community resulting in more research in the area in recent years (e.g., Human-Centric eXplainable AI in Education Workshop at 17th Educational Data Mining Conference\footnote{https://hexed-workshop.github.io/}). 

There are various tasks within the LA that focus on supporting learners and educators using various tools and methods. However, one of the most common objectives is the predictive modeling of learner success (with varying definitions of success), which focuses on the identification of the learners in need of help with their studies (Papamitsiou and Economides, 2014). Within the task of success prediction, the legacy learner and learning data are utilized for training the prediction model using the ML algorithm (Arnold and Pistilli, 2012; Waheed et al, 2020; Adnan et al., 2021). From the LA point of view, the prediction delivered by the ML model is used as a trigger for educational intervention. Thus the model itself is used as a tool by the lecturer, teaching assistant, or anyone responsible for supporting the students. Yet, there is a constant demand for providing not just the prediction itself, but also the “reasons behind the model decision” (Kuzilek et al., 2015). At this stage, again, the XAI comes into play and fosters the objectives of TLA (Drachsler H., 2018).

In the context of ML, the predictive models pursue the highest predictive accuracy, thus very often the so-called `black-box` models are preferred over the so-called `white-box` models, which in addition to the prediction provide intrinsically interpretable prediction (Guidotti et al., 2018; Biecek et al., 2021; Holzinger et al., 2022). However, to enable the power of XAI for the `black-box` models the post-hoc methods can be used (Pinto and Paquette, 2024). The XAI methods are primarily categorized into global and local. At the global level, they reveal which variables are important in the model. In contrast, at the local level, they answer questions about the contributions of variables in generating individual predictions (Molnar et al., 2020; Cavus et al., 2023). However, commonly used global and local tools, while sufficient for understanding the prediction made for a particular observation, are not sufficient for generating a counterfactual understanding of an undesirable outcome. Commonly used XAI methods (both local and global) are adequate for understanding particular observation predictions and not for generating a counterfactual understanding of an undesirable outcome (e. g. negative class in a binary classification problem).
To improve understanding of the undesirable outcome the method of counterfactual explanations (CEs) has become popular. CEs are defined as the minimal change in the variable values to flip the model's prediction into the intended outcome (Artelt and Hammer, 2019). In the frame of learner success prediction, the models may indicate an unfavorable outcome, but they do not provide recommendations to reverse the learner situation. Counterfactual explanations provide the extension of the baseline model and provide such a recommendation by highlighting necessary changes in the learner profile to reverse the negative outcome. Learners, teachers, and curriculum designers are guided toward actions or measures to be taken through their generated explanations.

The use of counterfactual explanations in LA has been explored in several studies (Tsiakmaki et al., 2021; Zhang et al., 2023; Afrin et al., 2023). All of the research works focused on providing a frame for delivering actionable insights to relevant stakeholders using the CE. Facing numerous counterfactual explanations due to the nature of optimization problems requires selecting those explanations that fulfill specific criteria beneficial for the stakeholder. Because of their background, challenges, and needs differences, each learner requires personalized counterfactual (Smith et al., 2022). 

The research presented in this paper focuses on using CE measures for the evaluation of the effect of balancing techniques used on the raw imbalanced dataset. More specifically we focus on the following research questions:\\

\noindent RQ1: What is the most appropriate method for generating the counterfactual explanations after balancing?\\

\noindent RQ2: How do balancing techniques affect the counterfactual explanations of student success prediction models?\\

This study compares the qualities of different counterfactual generation methods for students whose success prediction model developed after balancing the training dataset anticipates failing.  For the reproducibility of the developed approach, we used the Open University Learning Analytics Dataset (OULAD) (Kuzilek et al., 2017) as a raw data source. The study is essential in two ways: (1) because the missing evaluation of the counterfactual quality can lead to inefficient explanations, and this may compromise their trustworthiness (Artelt et al., 2021), (2) there is no uniformly better method for each domain (Dandl et al., 2023) and this is the first benchmark in the domain of LA, and (3) there are no many investigations on the effect of balancing methods on the counterfactual explanations (Gunonu et al, 2024).
The rest of the paper is organized using the following analysis approach. It examines the effect of balancing strategies on the quality of counterfactuals generated by the three most commonly used methods. Finally, the results are discussed.

\section{Methods}
\label{sec:methods}
This section contains details of the dataset, counterfactual explanations, resampling methods, and the experimental design.

\subsection{Data}
The OULAD dataset has been released by the research team of the Open University, the largest distance-learning institution in the UK. It is utilized to analyze the impact of the balancing strategies on the counterfactual generation methods. Typical course duration at the Open University is nine months and the course includes multiple assignments and a final exam. The most important assignments are Tutor Marked Assignments (TMAs), which represent critical milestones throughout the course. The dataset comprises learner demographics, assessment results, and click-stream log data from the interactions of learners with a Moodle-like Learning Management System (LMS).

For the analysis, the STEM course DDD and its 2013J and 2014J presentations studied by 3741 students have been selected. The course includes six TMAs. The final student result was used as the target variable for model training. Students with the result “Distinction” have been merged with students with the result “Pass”. Reducing the prediction task to binary classification to classes: “Pass” and “Fail”. We excluded the actively withdrawn students (n = 1328). The resulting dataset includes data from 2296 students. With the focus on actionable insights into the learner activity, we focused on the click-stream data only since it contains the most vital information (Kuzilek et al., 2015). Thus, the dataset consists of 42 predictors, which are numerical variables containing the weekly learner summary of online interactions with the LMS. Table~\ref{tab:variables} provides descriptions of the selected variables.

\begin{table}[]
    \centering
    \caption{The description of the variables used to train the student success prediction model}
    \label{tab:variables}
    \begin{tabular}{p{2cm}p{4.5cm}p{0.1cm}p{2.1cm}p{2cm}}\toprule
    Variable                & Description                                                       && Class         & Value      \\\midrule
    \texttt{final\_result}  & student’s final exam result                                       && categorical   & \{fail, pass\}   \\
    \texttt{week\_minus4}   & the number of clicks four  weeks before the course starts         && numeric       & $[0, 493]$  \\
    \texttt{week\_minus3}   & the number of clicks three weeks before the course starts         && numeric       & $[0, 765]$  \\
    \texttt{week\_minus2}   & the number of clicks two weeks before the course starts           && numeric       & $[0, 745]$  \\
    \texttt{week\_minus1}   & the number of clicks one week before the course starts            && numeric       & $[0, 987]$  \\
    \texttt{week\_0}        & the number of clicks before the course starts                     && numeric       & $[0, 1319]$  \\
    \texttt{week\_1}        & the number of clicks one week after the course starts             && numeric       & $[0, 525]$  \\
    ...                     & ...                                                               && ...           & ...  \\
    \texttt{week\_37}       & the number of clicks thirty-seven weeks after the course starts   && numeric       & $[0, 50]$  \\
    \bottomrule
    \end{tabular}
\end{table}
\subsection{Counterfactual explanations}
Counterfactual explanations (CE) illustrate "what-if" scenarios that emphasize the necessary alterations to the input data to change a model's output (Watcher et al., 2017). $X = [x_1, x_2, ..., x_p]$ represent a data matrix with $n$ observations and $p$ variables and $y$ be the response vector. The objective is to identify a function $f: X \rightarrow y$ that minimizes the expected value of the loss function $L$ in predictive modeling. A counterfactual $x^{\prime} \in \mathbf{R}^p$ of observation $x \in \mathbf{R}^p$ is determined by solving the following optimization problem:

\begin{equation}
    argmin_{x^{\prime} \in \mathbf{R}^p} L[f(x^{\prime}), y^{\prime}] + d(x, x^{\prime})
\end{equation}

\noindent where $\mathbf{R}^p$ represents $p$-dimensional real space, $L$ is a loss function that penalized the difference between the prediction $f(x^{\prime})$ and the desired outcome $y^{\prime}$, and $d$ is a distance function between the observation $x$ and $x^{\prime}$. A CE specifies the necessary adjustments in one or more variables to change the model's prediction. The distance function $d$ regulates the proximity between the original observation and the counterfactual. 

\begin{figure}
    \centering
    \caption{The counterfactual explanations for an observation}
    \label{fig:ce}
    \includegraphics[width=0.5\linewidth]{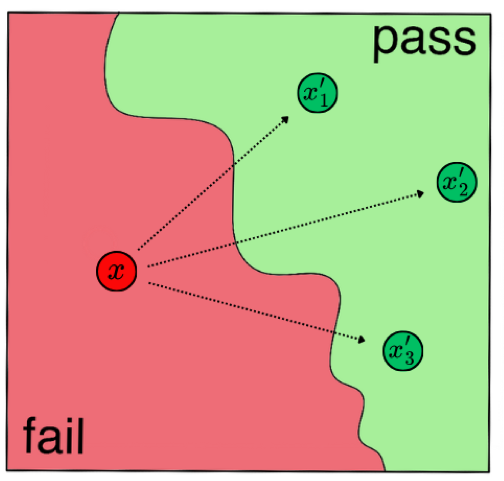}
\end{figure}

Figure~\ref{fig:ce} visualizes an observation and its counterfactuals. Assume that $f$ is a student success prediction model and x is a vector consists the variable values of a student. The prediction of the model $f$ for the student $x$ who has failed. The red zone shows the fail area, and the green one shows the pass area. They are divided by the decision boundary of the model. The CEs $x^{\prime}_1, x^{\prime}_2, x^{\prime}_3$ represent the ways how the student can pass. 

Counterfactuals strive to minimize the distance between the target observation and the counterfactual; however, additional properties are essential for a counterfactual explanation (Wachter et al., 2017; Karimi et al., 2020). Sparsity suggests altering the minimal number of variables to keep the explanation straightforward. Minimality aims for the most minor feasible changes in the variable values. Validity is ensured by reducing the difference between the counterfactual instance $x^{\prime}$ and the original observation $x$ while ensuring the model's output matches the desired label $y^{\prime}$. Proximity emphasizes the necessity for a slight variation between the factual and counterfactual features. Plausibility requires that counterfactual explanations remain realistic and closely follow the underlying data distribution. Over 120 known counterfactual generation methods; see Warren et al. (2023) for further details. However, we focused on three widely used counterfactual methods \textit{WhatIf Counterfactual Explanations}, \textit{Multi-Objective Counterfactual Explanations}, and \textit{Nearest Instance Counterfactual Explanations} to facilitate the comparison of counterfactual quality.\\

\noindent \textbf{What-if counterfactual explanations.} The What-if method (WhatIf) finds the observations closest to the observation x from the other observations in terms of Gower distance, solving the following optimization problem (Wexler et al., 2019):

\begin{equation}
    x^{\prime} \in argmin_{x \in X} d(x, x^{\prime}).
\end{equation}
\noindent \textbf{Multi-objective counterfactual explanations.} The Multiobjective Counterfactual Explanations (MOC) method aims to generate counterfactual explanations by optimizing multiple objectives simultaneously (Dandl et al., 2020). These objectives often include validity, proximity, sparsity, and plausibility. 

\begin{equation}
    x^{\prime} \in min_x \big( o_v(\hat{f}(x), y^{\prime}), o_p(x, x^{\prime}), o_s(x, x^{\prime}), o_{pl}(x, X) \big)
\end{equation}

\noindent where $o_v$, $o_p$, $o_s$, $o_{pl}$ are the objective functions for the desired properties validity, proximity, sparsity, and plausibility, respectively. Thus, it is expected that the counterfactuals generated by the MOC method are valid, proximity, sparse, and plausible. 
\noindent \textbf{Nearest instance counterfactual explanations.} The Nearest Instance Counterfactual Explanations (NICE) method identifies observations that are most similar to a given observation using the heterogeneous Euclidean overlap method (Burghmans et al., 2023). This approach allows for two options in the objective function, depending on the properties of proximity and sparsity, offering flexibility in how it can be applied.\\

The WhatIf method produces counterfactuals that are valid, proximal, and plausible. It has been demonstrated that the MOC method generates a higher number of counterfactuals that are closer to the training data and require fewer feature changes compared to other counterfactual methods (Dandl et al., 2020). Additionally, NICE specifically generates proximity-focused counterfactuals. However, no single method consistently outperforms others across datasets from various domains (Dandl et al., 2023). Therefore, evaluating the quality of the generated counterfactuals is essential, and we will conduct experiments to evaluate this in the following section.
\subsection{Balancing techniques}

The most commonly encountered challenge in designing predictive models with high discriminatory performance is an imbalanced class distribution in the response variable. In the binary case, the imbalance problem occurs when one class is observed less frequently. Models with such response variables tend to be biased toward the majority class in their predictions. Consequently, when dealing with the imbalance problem, models often have a significantly lower performance in correctly predicting the minority class than the majority class. In real-world problems, the class of interest is generally the minority class. For example, in predicting student dropouts, students who drop out are observed less frequently than those who do not. In the classification problem of predicting whether a student will complete a specific educational material or content module, students who do not complete the material are observed less frequently than those who do. In learning analytics, when considering student success prediction models, students who fail are observed less frequently than those who succeed. In these examples, students who drop out, do not complete educational materials, and fail constitute the minority class. Due to the nature of these problems, the focus is on identifying the minority class. The inaccurate models in correctly predicting the minority class is a problem that must be overcome in such scenarios. 

Solutions to this problem are divided into three categories: data-based, model-based, and weighting-based methods. The most commonly used data-based methods involve balancing class distributions through random undersampling or oversampling and synthetic data generation techniques. In undersampling, a subset of the majority class is randomly selected to match the minority class, whereas, in oversampling, the number of observations in the minority class is increased through resampling to match the size of the majority class (Chawla, 2010). In synthetic data generation methods, new observations are artificially generated from the distribution of the minority class to balance the size with the majority class (Elyan et al., 2021; Liu, 2022). Model-based methods are specific models developed to address the imbalance problem (Yin et al., 2020; Gu et al., 2022). Weighting-based methods aim to achieve higher prediction performance by penalizing the model more for errors in predicting the minority class (Zong et al., 2013; Tao et al., 2019). Although there are many methods to solve the classification problem in unbalanced data, in recent years, it has been found that these methods generally need to be revised and have adverse effects on classification models (Junior and Pisani, 2022; Stando et al., 2024; Cavus and Biecek, 2024; Carriero et al., 2024). These criticisms, mainly focusing on oversampling, undersampling, and synthetic data generation methods, brought the cost-sensitive approach to the fore (Gunonu et al., 2024). This study used data-based and weighting-based methods due to the mentioned criticism, their practical applications, and their frequent usage in the literature.

\subsection{Experimental design}

\begin{figure}[h]
    \centering
    \caption{The flow of the experiments}
    \label{fig:flow}
    \includegraphics[width=0.9\linewidth]{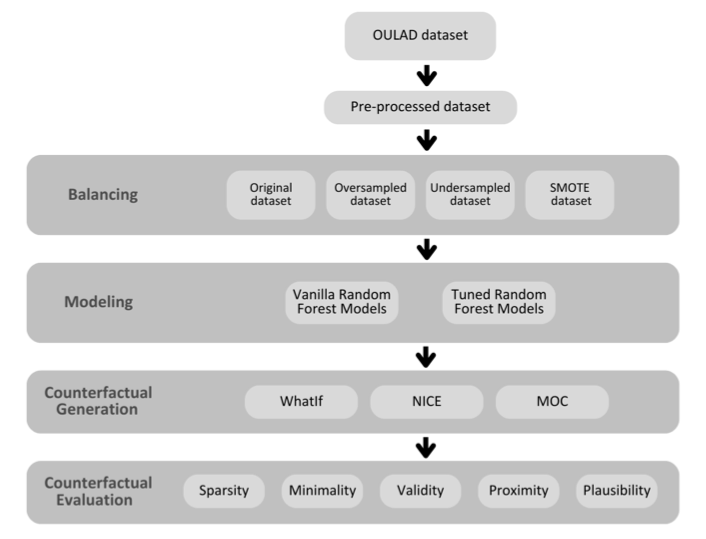}
\end{figure}

This paper focuses on which method provides the highest quality counterfactual explanations for the student success prediction model trained with and without hyperparameter tuning (i.e., vanilla model) regarding the imbalancedness problem using the OULAD dataset. Thus, the approach followed, which is given in Figure~\ref{fig:flow}, is (1) balancing the dataset, (2) training the model with and without hyperparameter tuning, (3) generating the counterfactuals, and (3) evaluating the effect of the balancing techniques of the imbalancedness problem producing the evaluation criteria.  

 \vspace{0.3cm}

\noindent \textbf{Balancing.} Two balancing strategies are used. The dataset is balanced using several resampling methods such as undersampling, oversampling, and SMOTE, and the models are trained on the original dataset with the cost-sensitive approach. \\

\noindent \textbf{Modeling.} The random forest algorithm is used in modeling because tree-based models exhibit lower prediction performance than alternative complex models in classifying tabular datasets (Grinsztajn et al., 2022). It is trained with and without hyperparameter tuning to achieve higher prediction performance. The performance of the random forest models trained on imbalanced (i.e., Original), balanced datasets by the oversampling, undersampling, SMOTE, and also trained with the cost-sensitive approach are compared. The costs are chosen as 2.37931 for the minority class (i.e., failed students) and 1 for the majority class regarding the imbalance ratio. Moreover, to achieve better predictive performance the models are tuned in terms of hyperparameters \texttt{mtry}, \texttt{splitrule}, and \texttt{min.node.size} using the 10-fold repeated cross-validation in addition to the vanilla versions of the model which is trained with the default values of the hyperparameters.\\

\noindent \textbf{Counterfactual generation.} After the modeling phase, the counterfactuals are generated for failing students which are estimated by the models using MOC, sparsity-based NICE (NICE\_sp), proximity-based NICE (NICE\_pr), and What-If methods. 
\section{Results \& Discussion}
In this section, the results are summarized. Firstly, the performance of the models is compared, and then the counterfactuals are evaluated to determine the best counterfactual generation method for the case considered in the paper.\\

\noindent \textbf{Model performance.} The performance of the random forest models trained on imbalanced and balanced datasets by the oversampling, undersampling, SMOTE, and cost-sensitive approach are given in Table~\ref{tab:perf}. The imbalance ratio of the test set is 2.41 (number of observations in the majority class/number of observations in the minority class), thus the performance evaluations should be using the F1 score as well as accuracy and AUC. 

\begin{table}[]
    \centering
    \caption{The performance of the random forest models on the test set over balancing strategies}
    \label{tab:perf}
    \begin{tabular}{p{2.4cm}p{1.5cm}p{1.5cm}p{1.5cm}p{1.5cm}p{1.5cm}p{1.5cm}}\toprule
                    & \multicolumn{3}{c}{Vanilla Random Forests Models}& \multicolumn{3}{c}{Tuned Random Forests Models}\\\cline{2-7}
                    & Accuracy  & AUC       & F1        & Accuracy  & AUC       & F1     \\\hline
    Original        & 0.8196    & 0.8549    & 0.7040    & 0.8044    & 0.8480    & 0.6450 \\
    Oversampling    & 0.8402    & 0.8652    & 0.6840    & 0.8366    & 0.8658    & 0.6795 \\
    Undersampling   & 0.7741    & 0.8552    & 0.6560    & 0.7812    & 0.8558    & 0.6611 \\
    SMOTE           & 0.8286    & 0.8620    & 0.6900    & 0.8321    & 0.8621    & 0.6907 \\
    Cost-sensitive  & 0.8357    & 0.8643    & 0.6940    & 0.8339    & 0.8671    & 0.6910 \\\bottomrule
    \end{tabular}
\end{table}

The vanilla Random Forest models generally outperform tuned models in terms of accuracy and F1 scores across most balancing strategies, particularly on original data and some resampling methods. Vanilla models demonstrate higher accuracy and more balanced F1 scores, especially under oversampling and SMOTE techniques. On the other hand, tuned models achieve slightly higher AUC values with cost-sensitive learning and SMOTE, indicating better classification discrimination. Sampling methods like oversampling and SMOTE improve performance for both vanilla and tuned models, while undersampling tends to decrease accuracy and F1 scores but maintains stable AUC values. Cost-sensitive learning offers balanced improvements, with both model types benefiting from enhanced AUC scores. Overall, while vanilla models excel in accuracy and F1 scores, tuned models show enhanced AUC values in specific conditions, highlighting the trade-offs between different performance metrics and modeling approaches. The tuned values of the hyperparameters for the models are given in Table~\ref{tab:hp} in the Appendix.\\

\noindent \textbf{Counterfactual evaluations.} The counterfactual generation methods can generate more than one explanation for an observation, also each method may generate different explanations. The number of counterfactuals generated by the methods is given in Table~\ref{tab:nce}. The MOC generates the highest number of counterfactuals independently from the balancing strategy and model while the NICE methods generate the lowest number of counterfactuals. The differences between the number of counterfactuals between the balancing strategies depend on the number of students that were predicted as failed by the models. The number of counterfactuals for the models is slightly different because of the difference between the models caused by the hyperparameter optimization.

\begin{table}[]
    \centering
    \caption{The number of counterfactuals generated by the methods across balancing strategies}
    \label{tab:nce}
    \begin{tabular}{p{1.5cm}p{1.5cm}rrrrr}\toprule
            & Model     & Original  & Undersampling & Oversampling  & SMOTE & Cost-sensitive    \\\midrule
    WI      & vanilla   & 2910      & 4050          & 2370          & 2890  & 2730  \\
            & tuned     & 2890      & 3950          & 2430          & 2800  & 2740 \\\midrule
    MOC     & vanilla   & 23321     & 28287         & 15934         & 24100 & 19570 \\
            & tuned     & 24932     & 38262         & 15997         & 23687 & 19401 \\\midrule
    NICE\_sp& vanilla   & 419       & 555           & 320           & 360   & 339 \\
            & tuned     & 390       & 530           & 327           & 336   & 530 \\\midrule
    NICE\_pr& vanilla   & 419       & 555           & 320           & 360   & 339 \\
            & tuned     & 390       & 530           & 327           & 336   & 530 \\\bottomrule
    \end{tabular}
\end{table}

\begin{figure}
    \centering
    \caption{Evaluation of counterfactual generation methods across tuning and balancing strategies}
    \label{fig:res}
    \includegraphics[width=1\linewidth]{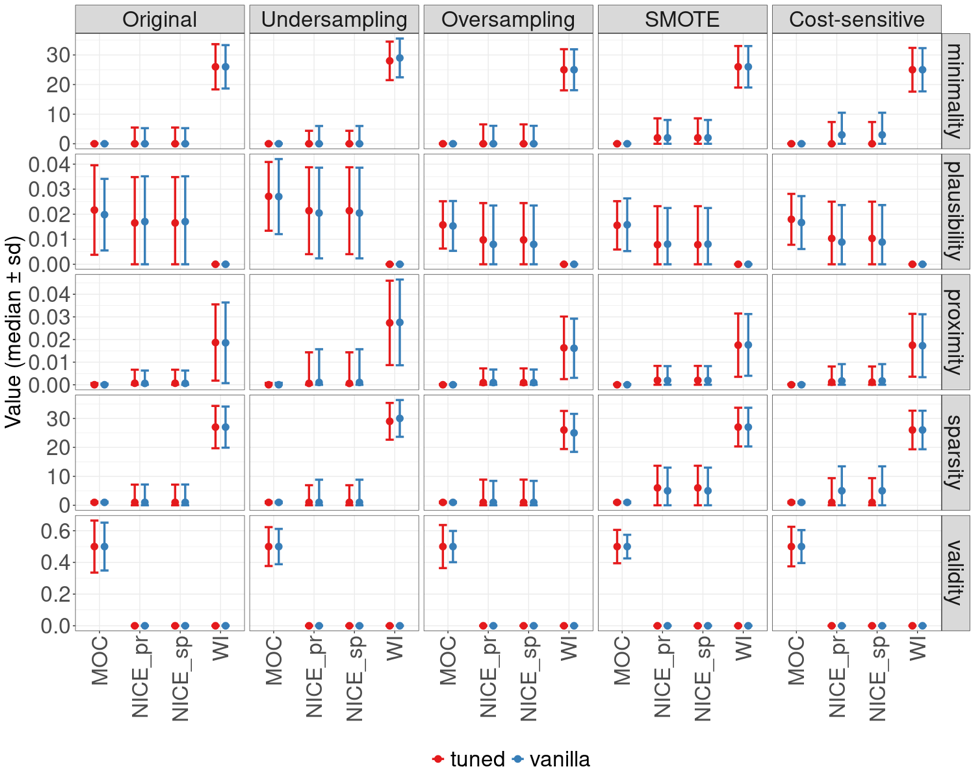}
\end{figure}

It is necessary to evaluate the quality or usefulness of the counterfactuals before deployment. Thus, we conduct a comparison study to analyze the effect of the conditions regarding the balancing and modeling strategies on the counterfactual quality. We aim to determine the best counterfactual generation method to find actionable insights from the student success prediction models trained on the OULAD dataset. The quality of counterfactuals is visualized using error bar plots as in Figure~\ref{fig:res}. An error bar plot shows the variability or uncertainty of data. It features error bars that extend above and below the median of the observations. Error bars can show measures of dispersion such as standard deviation, standard error, or confidence intervals, providing a visual indication of the reliability and precision of the data. Figure~\ref{fig:res} demonstrates that each method exhibits varying performance regarding quality metrics across different balancing and modeling strategies.

NICE\_sp and NICE\_pr consistently demonstrate superior performance with the models trained on the original dataset. The minimality and plausibility values are particularly low, with medians near 0 and minimal variability, suggesting robust performance. On the other hand, MOC and WI show much higher values, especially in minimality where median values reach around 30, indicating suboptimal outcomes. Similarly, in metrics like proximity and sparsity, NICE\_sp and NICE\_pr maintain low values, whereas MOC and WI exhibit considerably higher values, suggesting that these methods struggle with the original data distribution. 

When applying the Undersampling method, there is a general improvement in minimality values across all methods, though MOC and WI still trail behind NICE\_sp and NICE\_pr. While NICE\_sp and NICE\_pr continue to perform well with relatively low values across all metrics, the error bars suggest a slight increase in variability. MOC and WI, although showing some improvement, still exhibit higher plausibility and proximity values, indicating that undersampling does not fully mitigate their performance issues. 
The Oversampling method highlights the strengths of NICE\_sp and NICE\_pr even further. These methods maintain low values across all metrics, particularly in minimality and plausibility, where their performance remains nearly flawless with median values close to 0. In contrast, MOC and WI continue to struggle, showing higher values across metrics such as proximity and sparsity, with only marginal improvements compared to the Original and Undersampling strategies. This suggests that while oversampling enhances performance for NICE\_sp and NICE\_pr, it does not sufficiently benefit MOC and WI.

Moving to SMOTE, NICE\_sp, and NICE\_pr once again emerge as the top performers, maintaining low values across all metrics. The proximity and sparsity values for these methods remain minimal, reflecting strong and consistent performance. MOC and WI, however, continue to display higher values in metrics like minimality and validity, suggesting that even with synthetic data generation, these methods are less effective. The error bars for MOC and WI also indicate greater variability, reinforcing the idea that SMOTE does not significantly improve their robustness.

Finally, the cost-sensitive approach shows that NICE\_sp and NICE\_pr maintain their strong performance, with median values remaining low across all metrics. Particularly in minimality and plausibility, these methods exhibit near-perfect performance, with minimal error bars indicating consistent results. MOC and WI show slight reductions in their median values for some metrics, but they still lag significantly, with higher values in proximity and sparsity indicating ongoing performance issues. The consistent superiority of NICE\_sp and NICE\_pr across different balancing strategies, including Cost-sensitive approaches, underscores their robustness and reliability.

Tuned models consistently show improved performance compared to their vanilla counterparts across various balancing strategies. Tuned models trained on the original dataset exhibit lower minimality and plausibility values, indicating enhanced performance. In the Undersampling strategy, the gap between tuned and vanilla models narrows slightly, but tuned models still outperform vanilla ones. With Oversampling and SMOTE, the advantage of tuning becomes more pronounced, as tuned models maintain lower values across all metrics, while vanilla models show higher variability. The cost-sensitive approach further highlights the superiority of tuned models, particularly in minimality and validity, where they consistently demonstrate lower values and greater consistency. Overall, tuning leads to better and more reliable performance across different data conditions and metrics.

When focusing on RQ1: “What is the most appropriate method for generating counterfactual explanations after balancing?” the analysis highlighted the consistent superiority of NICE\_sp and NICE\_pr across various balancing strategies and metrics, demonstrating their robustness and reliability. To answer RQ2: “How do balancing techniques affect the counterfactual explanations of student success prediction models?” we find out that the impact of different data balancing strategies, such as SMOTE and the cost-sensitive approaches, further underscores the adaptability of these methods compared to MOC and WI, which generally underperform. Additionally, tuned models outperform their vanilla counterparts across all conditions, emphasizing the importance of model optimization in achieving optimal performance across diverse balancing strategies.
\section{Conclusion}
This study explored the impact of various balancing techniques on the generation of counterfactual explanations within student success prediction models. Our findings indicate that the choice of balancing strategy significantly influences the quality and characteristics of the counterfactuals generated by different methods, such as Multi-Objective Counterfactual Explanations (MOC), Nearest Instance Counterfactual Explanations (NICE), and WhatIf.\\

\noindent \textbf{Effectiveness of balancing techniques.} The results suggest that certain balancing techniques improve the validity and plausibility of counterfactuals, aligning them more closely with realistic scenarios that educators and students can act upon. For example, balancing methods that mitigate class imbalances not only enhanced the performance of the predictive models but also resulted in more actionable and sparse counterfactual explanations. These findings are consistent with previous research, which emphasizes the importance of balancing in training robust models for educational predictions (Artelt et al., 2021).\\

\noindent \textbf{Effect analysis of counterfactual generation methods.} Among the methods tested, MOC consistently produced counterfactuals that were closer to the original data distribution, showing a higher degree of plausibility and sparsity. This is particularly valuable in educational settings where changes to multiple variables might not be feasible. In contrast, the NICE method, which focuses on proximity, often generated explanations that were more straightforward but potentially less realistic. This trade-off highlights the need to select counterfactual generation methods based on the specific requirements of the educational context.\\

\noindent \textbf{Implications for Educational Interventions.} The insights gained from this study have significant implications for how educational institutions might use counterfactual explanations to inform interventions. By understanding how different balancing techniques affect the characteristics of counterfactuals, educators can better choose models and explanations that are not only accurate but also actionable and interpretable for students and staff.\\

This study contributes to the growing field of explainable artificial intelligence in education by demonstrating the critical role of balancing techniques in generating effective counterfactual explanations. These findings pave the way for more refined and targeted educational interventions, ultimately contributing to more personalized and supportive learning environments.

\section{Limitations and Future Work}

While this study provides a comprehensive analysis, some limitations warrant further investigation. The focus on a single dataset and specific counterfactual methods may limit the generalizability of the results. Future research should explore these effects across different datasets and additional counterfactual methods. Moreover, the long-term impact of using such explanations on student outcomes should be evaluated to better understand their practical utility in educational settings.
\section*{Acknowledgments}
The work in this paper is supported by the German Federal Ministry of Education and Research (BMBF), grant no. 16DHBKI045.

\section*{Supplemental Materials}

The materials for reproducing the experiments performed and the dataset are accessible in the following repository: \href{https://github.com/mcavs/JMEEP_paper}{https://github.com/mcavs/JMEEP\_paper}
\section*{References}

\noindent Adnan, M., Habib, A., Ashraf, J., Mussadiq, S., Raza, A. A., Abid, M., \& Khan, S. U. (2021). Predicting at-risk students at different percentages of course length for early intervention using machine learning models. IEEE Access, 9, 7519-7539.\\

\noindent Afrin, F., Hamilton, M., \& Thevathyan, C. (2023, June). Exploring Counterfactual Explanations for Predicting Student Success. In International Conference on Computational Science (pp. 413-420). Cham: Springer Nature Switzerland.\\

\noindent Arnold K. E. and Pistilli M. D. (2012). Course signals at Purdue: using learning analytics to increase student success. In Proceedings of the 2nd International Conference on Learning Analytics and Knowledge (LAK '12). Association for Computing Machinery, New York, NY, USA, 267–270.\\

\noindent Artelt, A., \& Hammer, B. (2019). On the computation of counterfactual explanations: A survey. arXiv preprint arXiv:1911.07749.\\

\noindent Artelt, A., Vaquet, V., Velioglu, R., Hinder, F., Brinkrolf, J., Schilling, M., \& Hammer, B. (2021). Evaluating the robustness of counterfactual explanations. In 2021 IEEE Symposium Series on Computational Intelligence (pp. 01-09). IEEE.\\

\noindent Biecek, P., \& Burzykowski, T. (2021). Explanatory model analysis: explore, explain, and examine predictive models. Chapman and Hall/CRC.\\

\noindent Brughmans, D., Leyman, P., \& Martens, D. (2023). Nice: an algorithm for nearest instance counterfactual explanations. Data mining and knowledge discovery, 1-39.\\

\noindent Carriero, A., Luijken, K., de Hond, A., Moons, K. G., van Calster, B., \& van Smeden, M. (2024). The harms of class imbalance corrections for machine learning based prediction models: a simulation study. arXiv preprint arXiv:2404.19494.\\

\noindent Cavus, M., \& Biecek, P. (2024). An Experimental Study on the Rashomon Effect of Balancing Methods in Imbalanced Classification. arXiv preprint arXiv 2405.01557.\\

\noindent Cavus, M., Stando, A., \& Biecek, P. (2023). Glocal Explanations of Expected Goal Models in Soccer. arXiv preprint arXiv:2308.15559.\\

\noindent Chawla, N. V. (2010). Data mining for imbalanced datasets: An overview. Data mining and knowledge discovery handbook, 875-886.\\

\noindent Dandl, S., Hofheinz, A., Binder, M., Bischl, B., \& Casalicchio, G. (2023). Counterfactuals: an R package for counterfactual explanation methods. arXiv preprint arXiv:2304.06569.\\

\noindent Dandl, S., Molnar, C., Binder, M., \& Bischl, B. (2020). Multi-objective counterfactual explanations. At the international conference on parallel problem solving from nature, Cham: Springer International Publishing, 448–469.\\

\noindent Drachsler, H. (2018). Trusted Learning Analytics. Synergie, 6, 40-43.\\

\noindent Elyan, E., Moreno-Garcia, C. F., \& Jayne, C. (2021). CDSMOTE: class decomposition and synthetic minority class oversampling technique for imbalanced-data classification. Neural computing and applications, 33, 2839-2851.\\

\noindent Grinsztajn, L., Oyallon, E., \& Varoquaux, G. (2022). Why do tree-based models still outperform deep learning on typical tabular data? Advances in neural information processing systems, 35, 507–520.\\

\noindent Guidotti, R. (2022). Counterfactual explanations and how to find them: literature review and benchmarking. Data Mining and Knowledge Discovery, 1-55.\\

\noindent Gunonu, S., Altun, G., \& Cavus, M. (2024). Explainable bank failure prediction models: Counterfactual explanations to reduce the failure risk. arXiv preprint arXiv:2407.11089.\\

\noindent Gu, Q., Tian, J., Li, X., \& Jiang, S. (2022). A novel Random Forest integrated model for imbalanced data classification problem. Knowledge-Based Systems, 250, 109050.\\

\noindent Hoel, T., Griffiths, D., \& Chen, W. (2017). The influence of data protection and privacy frameworks on the design of learning analytics systems. In Proceedings of the Seventh International Learning Analytics \& Knowledge Conference (pp. 243-252).\\

\noindent Holzinger, A., Saranti, A., Molnar, C., Biecek, P., \& Samek, W. (2022). Explainable AI methods-a brief overview. In International workshop on extending explainable AI beyond deep models and classifiers (pp. 13-38). Springer, Cham.\\

\noindent Junior, J. D. S. F., \& Pisani, P. H. (2022). Performance and model complexity on imbalanced datasets using resampling and cost-sensitive algorithms. In Fourth International Workshop on Learning with Imbalanced Domains: Theory and Applications (pp. 83-97). PMLR.\\

\noindent Karimi, A. H., Barthe, G., Balle, B., \& Valera, I. (2020). Model-agnostic counterfactual explanations for consequential decisions. At the International Conference on Artificial Intelligence and Statistics, Proceedings of Machine Learning Research, 895–905.\\

\noindent Kuzilek, J., Hlosta, M., \& Zdrahal, Z. (2017). Open university learning analytics dataset. Scientific data, 4(1), 1–8.\\

\noindent Kuzilek, J., Hlosta, M., Herrmannova, D., Zdrahal, Z., Vaclavek, J., \& Wolff, A. (2015). OU Analyse: analysing at-risk students at The Open University. Learning Analytics Review, 1-16.\\

\noindent Liu, J. (2022). Importance-SMOTE: a synthetic minority oversampling method for noisy imbalanced data. Soft Computing, 26(3), 1141-1163.\\

\noindent Molnar, C. (2020). Interpretable machine learning. Lulu.com.\\

\noindent Papamitsiou Z., and Economides A. (2014). Learning Analytics and Educational Data Mining in Practice: A Systematic Literature Review of Empirical Evidence. Journal of Educational Technology \& Society, 17(4), 49–64.\\

\noindent Pinto, J. D., \& Paquette, L. (2024). Towards a Unified Framework for Evaluating Explanations. arXiv preprint arXiv:2405.14016.\\

\noindent Siemens, G. and Baker, R. (2012). Learning analytics and educational data mining: towards communication and collaboration. In Proceedings of the 2nd International Conference on Learning Analytics and Knowledge (LAK '12). Association for Computing Machinery, New York, NY, USA, 252–254.\\

\noindent Smith, B. I., Chimedza, C., \& Bührmann, J. H. (2022). Individualized help for at-risk students using model-agnostic and counterfactual explanations. Education and Information Technologies, 1-20.\\

\noindent Stando, A., Cavus, M., \& Biecek, P. (2024, June). The effect of balancing methods on model behavior in imbalanced classification problems. In Fifth International Workshop on Learning with Imbalanced Domains: Theory and Applications (pp. 16-30). PMLR.\\

\noindent Tao, X., Li, Q., Guo, W., Ren, C., Li, C., Liu, R., \& Zou, J. (2019). Self-adaptive cost weights-based support vector machine cost-sensitive ensemble for imbalanced data classification. Information Sciences, 487, 31-56.\\

\noindent Tsiakmaki, M., \& Ragos, O. (2021). A Case Study of Interpretable Counterfactual Explanations for the Task of Predicting Student Academic Performance. In 2021 25th International Conference on Circuits, Systems, Communications and Computers (CSCC) (pp. 120-125). IEEE.\\

\noindent Wachter, S., Mittelstadt, B., \& Russell, C. (2017). Counterfactual explanations without opening the black box: Automated decisions and the GDPR. Harv. JL \& Tech., 31, 841–872.\\

\noindent Waheed, H., Hassan, S. U., Aljohani, N. R., Hardman, J., Alelyani, S., \& Nawaz, R. (2020). Predicting the academic performance of students from VLE big data using deep learning models. Computers in Human Behavior, 104, 106189.\\

\noindent Warren, G., Keane, M. T., Gueret, C., \& Delaney, E. (2023). Explaining groups of instances counterfactually for XAI: a use case, algorithm and user study for group-counterfactuals. arXiv preprint arXiv:2303.09297.\\

\noindent Wexler, J., Pushkarna, M., Bolukbasi, T., Wattenberg, M., Viégas, F., \& Wilson, J. (2019). The what-if tool: Interactive probing of machine learning models. IEEE transactions on visualization and computer graphics, 26(1), 56-65.\\

\noindent Yin, J., Gan, C., Zhao, K., Lin, X., Quan, Z., \& Wang, Z. J. (2020). A novel model for imbalanced data classification. In Proceedings of the AAAI Conference on Artificial Intelligence (Vol. 34, No. 04, pp. 6680-6687).\\

\noindent Zhang, H., Dong, J., Lv, C., Lin, Y., \& Bai, J. (2023). Visual analytics of potential dropout behavior patterns in online learning based on counterfactual explanation. Journal of Visualization, 26(3), 723-741.\\

\noindent Zong, W., Huang, G. B., \& Chen, Y. (2013). Weighted extreme learning machine for imbalance learning. Neurocomputing, 101, 229-242.

\newpage
\section*{Appendix}

\begin{table}[]
    \centering
    \caption{The best hyperparameter values of random forest models across balancing strategies}
    \label{tab:hp}
    \begin{tabular}{lccc}\toprule
                    & \texttt{mtry} & \texttt{splitrule}    & \texttt{min.node.size}   \\\midrule
    Original        & 41            & gini                  & 1  \\
    Oversampling    & 41            & extratrees            & 1  \\
    Undersampling   & 21            & gini                  & 1  \\
    SMOTE           & 41            & extratrees            & 1  \\
    Cost-sensitive  & 41            & extratrees            & 1  \\\bottomrule
    \end{tabular}
\end{table}

\end{document}